\documentclass[sigconf]{acmart}

\AtBeginDocument{%
  \providecommand\BibTeX{{%
    \normalfont B\kern-0.5em{\scshape i\kern-0.25em b}\kern-0.8em\TeX}}}

\setcopyright{acmlicensed}
\copyrightyear{2024}
\acmYear{2024}
\acmDOI{XXXXXXX.XXXXXXX}
\acmISBN{978-1-4503-XXXX-X/18/06}
\usepackage{amsmath,amsfonts}
\usepackage[linesnumbered,ruled,vlined]{algorithm2e}
\usepackage{pifont}
\usepackage{amsthm}
\usepackage{algpseudocode}
\usepackage{tcolorbox}
\usepackage{array}
\usepackage[caption=false,font=normalsize,labelfont=sf,textfont=sf]{subfig}
\usepackage{textcomp}
\usepackage{stfloats}
\usepackage{url}
\usepackage{verbatim}
\usepackage{graphicx}
\usepackage{ulem}
\usepackage{amsfonts}
\usepackage[utf8]{inputenc}
\usepackage{setspace}
\usepackage{cleveref}
\usepackage[utf8]{inputenc}
\crefname{section}{§}{§§}
\Crefname{section}{§}{§§}
\usepackage{caption}
\usepackage{booktabs}
\usepackage{makecell}
\usepackage{MnSymbol}
\usepackage{utfsym}
\usepackage{filecontents, pgffor}
\usepackage{balance}
\usepackage{latexsym}
\usepackage{color}
\usepackage[longtable]{multirow}
\usepackage{longtable}
\usepackage{listings, xcolor}
\definecolor{verylightgray}{rgb}{.97,.97,.97}

\lstdefinelanguage{Solidity}{
	keywords=[1]{anonymous, assembly, assert, balance, break, call, callcode, case, catch, class, constant, continue, constructor, contract, debugger, default, delegatecall, delete, do, else, emit, event, experimental, export, external, false, finally, for, function, gas, if, implements, import, in, indexed, instanceof, interface, internal, is, length, library, log0, log1, log2, log3, log4, memory, modifier, new, payable, pragma, private, protected, public, pure, push, require, return, returns, revert, selfdestruct, send, solidity, storage, struct, suicide, super, switch, then, this, throw, transfer, true, try, typeof, using, value, view, while, with, addmod, ecrecover, keccak256, mulmod, ripemd160, sha256, sha3}, % generic keywords including crypto operations
	keywordstyle=[1]\color{blue}\bfseries,
	keywords=[2]{address, bool, byte, bytes, bytes1, bytes2, bytes3, bytes4, bytes5, bytes6, bytes7, bytes8, bytes9, bytes10, bytes11, bytes12, bytes13, bytes14, bytes15, bytes16, bytes17, bytes18, bytes19, bytes20, bytes21, bytes22, bytes23, bytes24, bytes25, bytes26, bytes27, bytes28, bytes29, bytes30, bytes31, bytes32, enum, int, int8, int16, int24, int32, int40, int48, int56, int64, int72, int80, int88, int96, int104, int112, int120, int128, int136, int144, int152, int160, int168, int176, int184, int192, int200, int208, int216, int224, int232, int240, int248, int256, mapping, string, uint, uint8, uint16, uint24, uint32, uint40, uint48, uint56, uint64, uint72, uint80, uint88, uint96, uint104, uint112, uint120, uint128, uint136, uint144, uint152, uint160, uint168, uint176, uint184, uint192, uint200, uint208, uint216, uint224, uint232, uint240, uint248, uint256, var, void, ether, finney, szabo, wei, days, hours, minutes, seconds, weeks, years},	% types; money and time units
	keywordstyle=[2]\color{teal}\bfseries,
	keywords=[3]{block, blockhash, coinbase, difficulty, gaslimit, number, timestamp, msg, data, gas, sender, sig, value, now, tx, gasprice, origin},	% environment variables
	keywordstyle=[3]\color{violet}\bfseries,
	identifierstyle=\color{black},
	sensitive=true,
	comment=[l]{//},
	morecomment=[s]{/*}{*/},
	commentstyle=\color{gray}\ttfamily,
	stringstyle=\color{red}\ttfamily,
	morestring=[b]',
	morestring=[b]"
}

\begin{document}
% \title{ABI-Aware Bytecode-Level Vulnerability Detection for Ethereum Smart Contracts}

\title{Towards Understanding Deep Learning Model in Image Recognition via Coverage Test}

% \balance
\author{Wenkai Li}

\affiliation{%
  \institution{Hainan University}
  \city{Haikou}
  \country{China}
}
\email{cswkli@hainanu.edu.cn}

\author{Xiaoqi Li}
% \authornote{The corresponding author}
\affiliation{%
  \institution{Hainan University}
  \city{Haikou}
  \country{China}
}
\email{csxqli@ieee.org}

\author{Yingjie Mao}
\affiliation{%
  \institution{Hainan University}
  \city{Haikou}
  \country{China}
}
\email{yingjiemao@hainanu.edu.cn}

\author{Yishun Wang}
\affiliation{%
  \institution{Hainan University}
  \city{Haikou}
  \country{China}
}
\email{yishunwang@hainanu.edu.cn}

% \date{February 2023}

\renewcommand{\shortauthors}{Wenkai and Xiaoqi, et al.}

\begin{abstract}
Deep neural networks (DNNs) play a crucial role in the field of artificial intelligence, and their security-related testing has been a prominent research focus. By inputting test cases, the behavior of models is examined for anomalies, and coverage metrics are utilized to determine the extent of neurons covered by these test cases. With the widespread application and advancement of DNNs, different types of neural behaviors have garnered attention, leading to the emergence of various coverage metrics for neural networks. However, there is currently a lack of empirical research on these coverage metrics, specifically in analyzing the relationships and patterns between model depth, configuration information, and neural network coverage. This paper aims to investigate the relationships and patterns of four coverage metrics: primary functionality, boundary, hierarchy, and structural coverage. A series of empirical experiments were conducted, selecting LeNet, VGG, and ResNet as different DNN architectures, along with 10 models of varying depths ranging from 5 to 54 layers, to compare and study the relationships between different depths, configuration information, and various neural network coverage metrics. Additionally, an investigation was carried out on the relationships between modified decision/condition coverage and dataset size. Finally, three potential future directions are proposed to further contribute to the security testing of DNN Models.
\end{abstract}


\maketitle

\section{Introduction}
\label{sec:intro}

Deep Neural Networks (DNNs) play a significant role in the development of deep learning technologies and have been applied to a wide array of specific tasks, including image segmentation, object detection, and pose estimation \cite{1}. A DNN is composed of multiple layers, including input, output, and hidden layers. Compared to shallow neural networks, DNNs construct higher-level features of data by increasing the number of layers. They learn the feature representation of input data in the hidden space by updating the weight information in the neurons through forward or backward propagation. With large-scale AI models being applied to various security-related downstream tasks, such as autonomous vehicle systems and malicious software detection systems \cite{1}, DNNs are often used as specific basic functional components \cite{3}.

\begin{figure}
    \centering
    \includegraphics[width=0.85\linewidth]{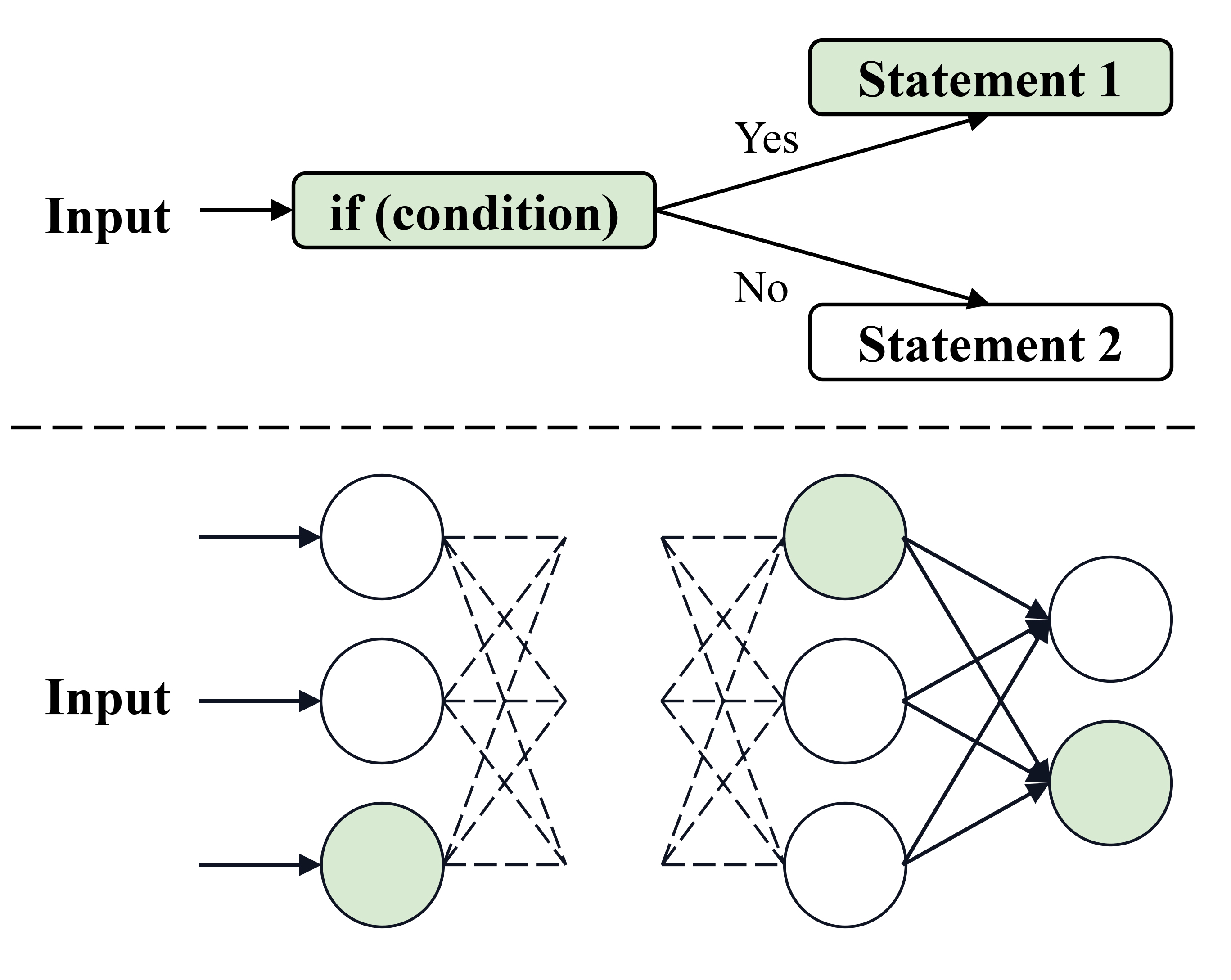}
    \vspace{-3ex}
    \caption{Comparison of traditional code and deep neural network program flow. Above the dotted line is the traditional code flow diagram; below the dotted line is the deep neural network execution diagram. Green indicates the base cell covered by the input example.}
    \label{fig:motivation}
    \vspace{-4ex}
\end{figure}

Due to their complex network structure, DNNs may produce different judgments for inputs with minor differences, leading to erroneous results~\cite{ghaleb2023achecker,tsankov2018securify,tikhomirov2018smartcheck}. Therefore, testing DNNs to improve system quality has become extremely important. Existing work uses traditional software testing methods to generate test cases to check for potential vulnerabilities \cite{4}. To evaluate the effectiveness of these generated test cases, various coverage metrics have been proposed, such as neuron coverage \cite{4} and Modified Condition/Decision Coverage (MC/DC) \cite{5}. Also, to focus on the behavior of neurons in different locations within the network structure, various coverage metrics have different characteristics \cite{6}. For example, as shown in Figure \ref{fig:motivation}, statement coverage mainly checks the executable statements involved in the test cases; neuron coverage evaluates the proportion of neurons covered by the test cases.

Many works have focused on the priority of testing inputs \cite{7} to guide security personnel to achieve detection effects in a shorter time, thereby saving resource consumption during testing~\cite{bhargavan2016formal,FerreiraEtAl2020ASE}. Therefore, current empirical research focuses on comparing test case priority indicators \cite{7} with gradient-based attack methods \cite{8} for effectiveness. These methods aim to select more suitable evaluation metrics to generate test cases to evaluate model quality \cite{10}. However, these studies on neuron coverage only focus on test cases and lack research that considers the DNN model itself as an influencing factor~\cite{zhong2024prettysmart,white2016deep,sajnani2016sourcerercc,tian2022ethereum}. Nowadays, many excellent DNN structures have been proposed, such as LeNet, AlexNet, and VGG \cite{1}. DNNs with different structures not only have different model structures but also different model depths. As a model component in AI models, the variation of DNNs is an important factor to be considered \cite{11}. 

Therefore, to explore the relationship between model depth and various coverage metrics, this paper conducts an empirical study, treating various deep neural networks as black boxes, and testing their relationship with multiple white box coverage metrics. By controlling four variables - the dataset, model structure, model depth, and coverage configuration parameters - through quantitative analysis methods, we analyze the impact of depth changes in deep neural networks of different structures on coverage metrics. In this series of experiments, we selected three different structures of deep neural networks, namely LeNet, VGG, and ResNet, and constructed 10 different deep neural networks ranging from 5 to 54 layers within them. Subsequently, by training each model in the MNIST and CIFAR10 datasets, and using the test set as test case inputs to the model, we analyze each coverage metric.

\textbf{Contributions.} The main contributions are as follows:
% \vspace{-1ex}
\begin{itemize}
    \item We selected 10 different depth models in the LeNet, VGG, and ResNet neural network structures, respectively, and summarized the rules of their neural network coverage.
    \item We studied the neuron coverage metrics at different levels to empirically analyze the rules of various coverage metrics under various configuration parameter conditions.
    \item We conducted LeNet network experiments based on MNIST to analyze the rules of the traditional coverage metric MD/DC, as well as its relationship with neuron coverage. 
\end{itemize}

% \textbf{Paper Organizations.} 
% The remainder of the paper is organized as follows. Section \ref{sec:background} provides the background and motivation of the paper and Section \ref{sec::method} details the implementation of \textsc{DeFiTail}. In Section \ref{sec::evaluation}, we show the experimental results to demonstrate the effectiveness of our proposed method. Finally, we review related literature in Section \ref{sec::relatedwork} and conclude our work in Section \ref{sec::conclusion}.

% The limitations of \textsc{COBRA} and possible future optimizations are discussed in Section \ref{sec:lim}. 

% \vspace{-3ex}

\section{Background}
\label{sec:background}
In this section, we mainly introduce the coverage metrics of related deep neural networks, including neural network coverage and traditional coverage metrics.

\subsection{Neuron Network Coverage}
In order to make structured coverage metrics more suitable for deep neural networks, a variety of neural network coverage metrics have been proposed to guide deep neural networks~\cite{he2020characterizing,ghaleb2020effective,neamtiu2005understanding,yang2024uncover}. The operation of software depends on the execution continuity. For example, traditional software depends on the previous statement, while deep neural networks depend on the previous layer of neurons~\cite{xiao2025wakemint,wang2021non,qian2022smart,chu2023survey}. As shown in Figure \ref{fig:motivation}, the traditional program has branches. After the program runs, the test case input can determine whether there is a problem in the code corresponding to the program statement~\cite{nadini2021mapping,ante2023non,tolmach2021survey,gao2020checking}. Therefore, the more code statements the traditional program covers, the higher the probability of finding code problems. For deep neural networks, the coverage should be based on the neuron activity to determine whether the input has passed through all possible cases~\cite{huang2021hunting,ma2024combining,szabo1996smart,ante2022non}.

According to the strength of neuronal activity, neural network coverage can be divided into main functional area coverage and boundary area coverage~\cite{mikolov2013efficient,choromanski2020rethinking,beltagy2020longformer}. Moreover, due to the different granularity features in different layers of deep neural networks \cite{12}, the layer coverage index is also introduced in this paper.

\subsection{Main Functional Area Coverage Indicators}

\noindent\textbf{Neuron Coverage.}
Neuron Coverage (NC) \cite{14} refers to the ratio of the number of neurons whose activation values exceed a threshold when the neural network is input with a test input set to the total number of all neurons~\cite{srinivasa2020fast,malkov2018efficient,ain2019systematic}. The formula for NC can be expressed as $NCov = \frac{|\{ n | \forall i \in T, O( n, i ) > t \}|}{|N|},$ where $n$ represents a neuron, and all neurons constitute the set $N$; $i$ represents the test input, and all test inputs constitute the set $T$; $O()$ represents the activation value of a neuron, $O(n, i)$ means the activation result after inputting $i$ on neuron $n$; $t$ is the threshold for a neuron to be activated.

\noindent\textbf{K-multisection Neuron Coverage.} K-multisection Neuron Coverage (KMNC) \cite{4} refers to dividing the activation value domain of all neurons into k equal parts, and the ratio of the part covered by the test input set $T=\{x_1, x_2, \dots\}$ to the total part when inputting the neural network~\cite{liang2024identity,zhu2022bytecode,zhang2024combining,pasqua2023enhancing}. The activation value domain is obtained by inputting the training dataset into the model and observing the activation data statistics of all neurons during the training process. The formula for KMNC can be expressed as $\frac{\sum | \{ S^n_j | \exists x \in T; C(x, n) \in S^n_j \}_{n \in N} |}{k * |N|},$ where $C(x, n)$ represents the activation value of input instance $x$ in neuron $n$; $j$ is a value from 1 to $k$; $\frac{| \{ S^n_j | \exists x \in T; C(x, n) \in S^n_j \}|}{k}$ represents the KMNC of neuron $n$.

\noindent\textbf{Neuron Boundary Coverage.} The KMNC metric is applied to the activation value range $[L, H]$, i.e., it can only cover $C(x, n)$ in the activation value range~\cite{bojanowski2017enriching,zhang2024acfix}. Therefore, the Neuron Boundary Coverage (NBC) metric \cite{4} is defined as the ratio of the number of neurons covered outside the activation value range $(-\infty, L) \cup (H, +\infty)$ to the total number of neurons. The formula for the NBC metric can be represented as $\frac{|\{n | \exists x \in T; C(x, n) \in (-\infty, L)\}| + |\{n | \exists x \in T; C(x, n) \in (H, +\infty)\}|}{2 \times |N|},$ where $n$ is a neuron in $N$ \cite{7}, and the number of neurons that exist in the lower domain $(-\infty, L)$ and the upper domain $(H, +\infty)$ are both $|N|$. In detail, we denote the size of each region as $\tau = (H - L) / k$.

\noindent\textbf{Strong Neuron Activation Coverage.} Consistent with NBC, Strong Neuron Activation Coverage (SNAC) \cite{4, 13} is also used to detect neuron coverage outside the activation value range. However, to analyze the behavior of highly active neurons in the model, SNAC only considers neurons with activation values in the upper domain, that is, the neuron activation value is within $(H, +\infty)$. In addition, its formula with $T$ as the input case set can be represented as $\frac{|\{n | \exists x \in T; C(x, n) \in (H, +\infty)\}|}{|N|}.$

\subsection{Layer Coverage Indicator}

\noindent\textbf{Top-k Neuron Coverage.} Neuron coverage calculates the proportion of the number of neurons covered by test cases to the total number of neurons \cite{7}, while the layer coverage metric calculates the coverage rate from the layer granularity. Top-k Neuron Coverage (TopkNC) \cite{4,13} refers to the number of times the neurons with the highest activation scores belong to the top k in their layer, that is, the number of the top k neurons on each layer as a proportion of the total number of neurons. The formula for the TopkNC metric can be expressed as $\frac{|\sum_{x \in T} (\sum_{i=1}^{l} topk(x, i))|}{|N|},$ where $l$ represents the number of all layers; $topk(x, i)$ represents the top $k$ neurons with the highest activation scores in the $i$th layer when $x$ is the input test case.

\subsection{Traditional Coverage Indicator}
\noindent\textbf{Modified Condition/Decision Coverage.} Modified Condition / Decision Coverage (MC/DC) \cite{15} requires both condition and decision coverage, and each decision condition can be influenced by any condition. The core idea is that, for each compound condition, the atomic conditions within it contribute to some positive or negative result of the compound condition, i.e., when the atomic condition changes, the result of the compound condition also changes correspondingly. Consequently, in deep neural networks \cite{16}, if a neuron $\sigma$ at layer $l$ is taken as a decision, then the conditions refer to all neurons at layer $l-1$ that can influence the behavior of neuron $\sigma$.

\section{EXPERIMENTS}
\label{sec::method}
In this section, we mainly introduce the main experimental procedure. First, three key questions related to neuron coverage are posed. We then conducted a series of experiments to answer the following questions.

\noindent RQ 1: \textbf{How do model depths impact the network coverage?}\\
\noindent RQ 2: \textbf{How do the parameters impact the network coverage?}\\
\noindent RQ 3: \textbf{Are traditional coverage metrics effective in networks?}

\subsection{Experimental Setup}
In this paper, we perform experiments on deep neural networks with three different network structures, and there are a total of 10 different numbers of layers. We select LeNet\cite{17} to represent small-scale deep models, and VGG\cite{18} and ResNet\cite{19} to represent large-scale deep models. In addition, VGG uses a sequential structure, while ResNet uses a non-sequential structure. We evaluate the LeNet network structure using the MNIST dataset, and evaluate the VGG and ResNet network structures using the CIFAR10 dataset. According to the description in Table \ref{tab:dataset_details}, the MNIST dataset consists of $28 \times 28 \times 1$ pixel size grayscale images of handwritten digits, with 60,000 images for training and 10,000 images for testing. The CIFAR10 dataset contains color images of $32 \times 32 \times 3$ pixel size, with a total of 10 categories, each containing 6,000 images. In our evaluation process, we randomly select 50,000 images from the CIFAR10 dataset for training, and use the remaining part for testing coverage metrics. All experiment results are completed on a Ubuntu Server 22.04 device equipped with an Intel(R) Core(TM) i9-13900k CPU and NVIDIA GeForce RTX 4070 Ti.

\begin{table}[ht]
    \centering
    \caption{The Statistical Information of Datasets}
    \label{tab:dataset_details}
    \vspace{-2ex}
    \begin{tabular}{c c c c}
    \toprule
         Dataset & Pixel Size & Class Size & Training : Testing \\
    \midrule
         MNIST &  $28 \times 28 \times 1$  & 10  &  60K : 10k\\
    \hline
         CIFAR10 & $32 \times 32 \times 3$ & 10  &  50K : 10k\\
    \bottomrule
    \end{tabular}
\end{table}

% \subsection{The Details of Experiment}

\begin{table*}[ht]
\caption{Neuron Coverage and K-Multisection Neuron Coverage for Deep Models}
    \vspace{-2ex}
    \centering
    \label{tab:NC_KMulti}
    \begin{tabular}{ll|ccc|cccc|cccc}
    \toprule
\multirow{3}{*}{Indicators} & \multirow{3}{*}{Parameters} & \multicolumn{9}{c}{Models} & \\*
\cline{3-13}
    &    & \multicolumn{3}{c}{LeNet} & \multicolumn{4}{c}{VGG} & \multicolumn{4}{c}{ResNet} \\*
\cline{3-5}
\cline{6-9}
\cline{10-13}
    &    & 5 layers     & 6 layers     & 7 layers & 11 layers    & 13 layers    & 16 layers    & 19 layers    & 21 layers    & 37 layers    & \multicolumn{2}{c}{54 layers}     \\*
\midrule
\multirow{5}{*}{NC}      & $\tau$=0.3                & 0.5476 & 0.7391 & 0.8256  & 0.8758 & 0.8728 & 0.885  & 0.8833 & 0.9994 & 0.9944 & \multicolumn{2}{l}{0.9995}  \\*
                             & $\tau$=0.45               & 0.3571 & 0.6594 & 0.7713  & 0.8246 & 0.8196 & 0.8315 & 0.7713 & 0.9726 & 0.9507 & \multicolumn{2}{l}{0.9795}  \\*
                             & $\tau$=0.6                & 0.2619 & 0.6014 & 0.7054  & 0.7098 & 0.7008 & 0.6908 & 0.6612 & 0.9015 & 0.8789 & \multicolumn{2}{l}{0.8378}  \\*
                             & $\tau$=0.75               & 0.2381 & 0.5942 & 0.6705  & 0.4823 & 0.4667 & 0.4578 & 0.4152 & 0.7536 & 0.7176 & \multicolumn{2}{l}{0.5888}  \\*
                             & $\tau$=0.9                & 0.2381 & 0.5797 & 0.6434  & 0.2128 & 0.208  & 0.196  & 0.1709 & 0.4314 & 0.3671 & \multicolumn{2}{l}{0.2036}  \\* 
\midrule
\multirow{5}{*}{KMNC} & $k$=10                & 0.95   & 0.8529 & 0.907   & 0.983  & 0.983  & 0.9811 & 0.9774 & 0.9961 & 0.9958 & \multicolumn{2}{l}{0.9945}  \\*
                             & $k$=100               & 0.8267 & 0.7273 & 0.7824  & 0.956  & 0.9542 & 0.9605 & 0.9556 & 0.9555 & 0.9622 & \multicolumn{2}{l}{0.9582}  \\*
                             & $k$=1000              & 0.6023 & 0.5433 & 0.591   & 0.8831 & 0.8753 & 0.8547 & 0.8618 & 0.8246 & 0.8122 & \multicolumn{2}{l}{0.8201}  \\*
                             & $k$=5000              & 0.4464 & 0.3493 & 0.3559  & 0.7673 & 0.7278 & 0.6414 & 0.6528 & 0.6634 & 0.6502 & \multicolumn{2}{l}{0.6545}  \\*
                             & $k$=10000             & 0.3536 & 0.2506 & 0.2446  & 0.6814 & 0.6848 & 0.5407 & 0.5521 & 0.5832 & 0.5774 & \multicolumn{2}{l}{0.5765}  \\
\bottomrule
    \end{tabular}
    \vspace{-2ex}
\end{table*}

\subsection{The Neuron Coverage}
To explore the influence of different model structures and depths on neuron coverage, we selected three deep models with different structures: LeNet, VGG, and ResNet. Among them, LeNet and VGG adopt a sequential deep network structure, while ResNet adopts a non-sequential deep network structure. In this section, we conduct experiments, which cover models with different depths from 5 to 54 layers, and evaluate the coverage of the neuron coverage metrics at different thresholds.
For the small-scale deep model LeNet, the difference between different layers is the number of final fully connected layers. For example, LeNet1 is a sequential deep network architecture with five layers, in order: convolutional layer, average pooling layer, convolutional layer, average pooling layer, and fully connected layer; LeNet4 is a 6-layer structure, and according to the sequential network hierarchy, a layer of fully connected gap exists between LeNet1 and LeNet4; While LeNet5 is a 7-layer structure, its sequential network layer level design has more number of fully connected layers, and this increase number is one.

\begin{figure}[ht]
\vspace{-2ex}
    \centering
    \subfloat[NC]{\includegraphics[width=.45\columnwidth]{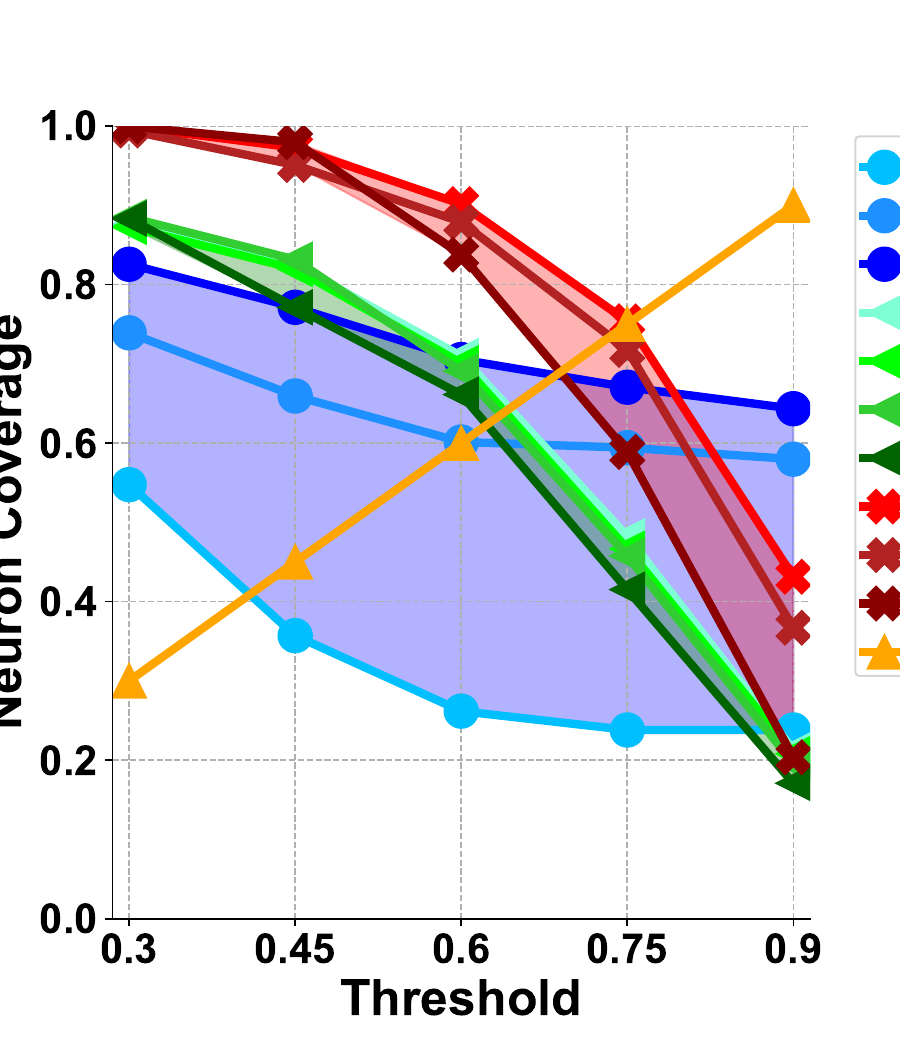}}
    \subfloat[KMNC]{\includegraphics[width=.45\columnwidth]{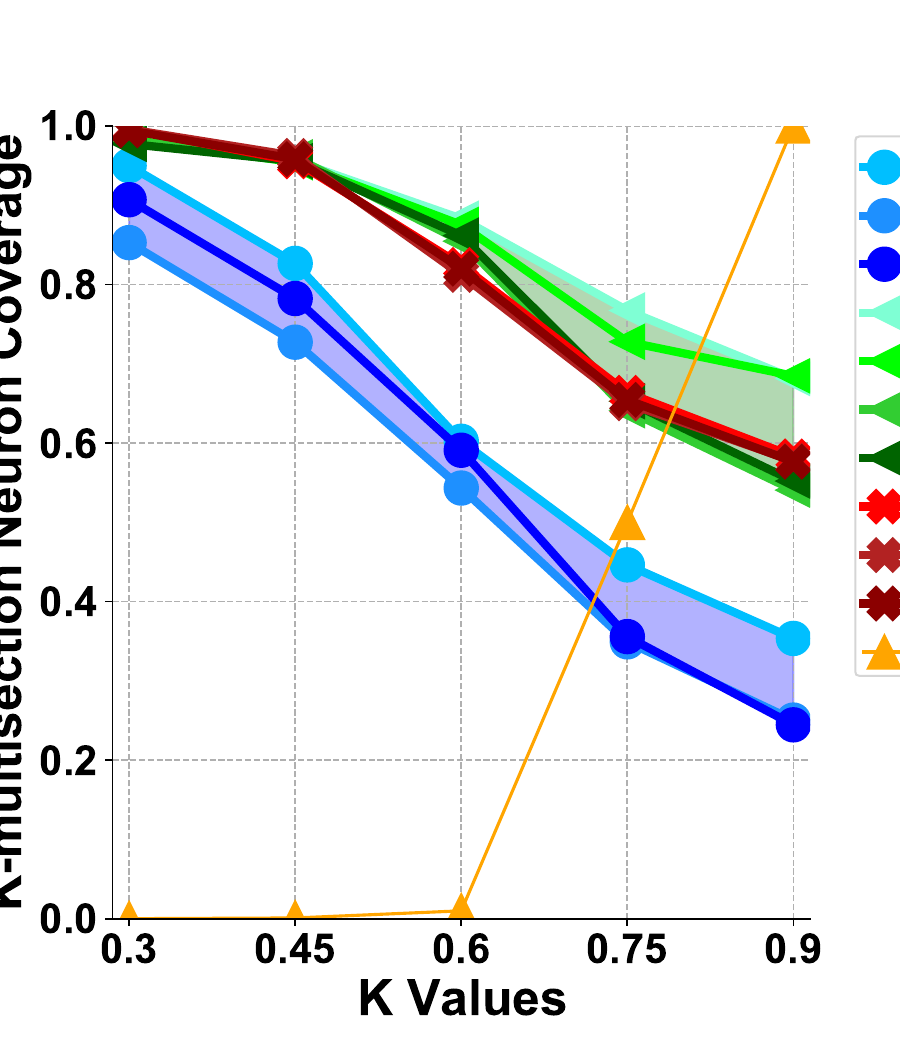}}
    \vspace{-2ex}
    \caption{The NC and KMNC in Different Models. Blue groups represent the coverages of LeNets, green groups for VGGs, and red groups for ResNets. }
    \label{fig:Neuron_Coverages}
    
\end{figure}

However, as shown in Figure \ref{fig:Neuron_Coverages}, for the relatively small-scale sequential network structure LeNet, the neuron coverage has an inverse relationship with the threshold, and the rate of neuron coverage decline has an inverse relationship with the number of threshold increases. According to our experimental results, the neuron coverage of the LeNet model structure will tend to a stable value $\delta$ as the threshold increases. Specifically, the stable value $\delta$ corresponding to LeNet1 is about 0.24, the stable value $\delta$ corresponding to LeNet4 is about 0.58, and the stable value $\delta$ corresponding to LeNet5 is about 0.64. Moreover, as depth increases, LeNet models with greater depth will have higher neuron coverage. Therefore, compared to the neuron coverage gap between LeNet4 and LeNet5, the gap between LeNet1 and LeNet4 increases with the enhancement of the threshold.

The VGG model consists of a sequential deep network structure composed of five convolutional segments, and different from the way of depth increase of the LeNet model, VGG achieves different depths by adding convolutional layers in different convolutional segments. In this experiment, we tested these deep neural network architectures VGG11 (11 layers), VGG13 (13 layers), VGG16 (16 layers), and VGG19 (19 layers), where the number of layers refers to the number of convolutional and fully connected layers. As shown in Figure \ref{fig:Neuron_Coverages}, at lower thresholds, the neuron coverage metric is difficult to distinguish VGG models with different depths, and the gap between the neuron coverage metrics is not large in the distribution of the individual thresholds. However, according to the results in Table \ref{tab:NC_KMulti}, when the threshold is larger than 0.6, the neuron coverage decreases as the depth of the VGG model increases. In addition, we observe that the rate of decline of the neuron coverage metric increases with the threshold, independent of the depth of the model. Therefore, the decreasing trend of neuron coverage in the VGG model is directly opposite to that in the LeNet model.

The ResNet model consists of a network structure consisting of five convolutional segments. However, unlike the VGG model, ResNet employs residual unit connections and is therefore a non-sequential structure. In Table \ref{tab:NC_KMulti}, the number of layers represents the total number of convolutional, pooling, and fully connected layers. We explored the relationship between ResNet neural networks with non-sequential structures, such as ResNet18 (21 layers), ResNet34 (37 layers), and ResNet50 (54 layers), and the neuron coverage index. As shown in Figure \ref{fig:Neuron_Coverages}, the neuron coverage is proportional to the model depth at threshold values above 0.6. At thresholds 0.3 and 0.45, ResNet34 has lower neuron coverage than ResNet18 and ResNet50 networks. In addition, the decline rate law of neuron coverage is basically consistent with the VGG model, which increases with the increase of the threshold. However, compared with the VGG model, the neuron coverage of ResNet decreases more obviously, and the rate of decline increases with the depth.

\noindent\textbf{Answer to RQ1.}
As the model depth increases, the neuron coverage decreases. In addition, since the rate of decline of neuron coverage is related to the model depth as well as the threshold range, the answers to Question one and two are relevant. When the configuration parameter of this coverage metric, namely the threshold, is greater than or equal to 0.6, the rate of decline of LeNet and ResNet models will accelerate as the depth of the model increases. However, when the threshold is 0.3 or 0.45, the decline rate of LeNet and ResNet models also slows down as the model depth increases. In addition, there is a weak correlation between the rate of descent and the depth of the model for the VGG model.

\subsection{The K-Multisection Neuron Coverage}
The KMNC aims to focus on the main functional regions involved in the trained model to obtain the average coverage of K neurons in the main functional regions of the test set.
For the small-scale LeNet model, as shown in Table \ref{tab:NC_KMulti}, the highest neuron coverage of the LeNet model is 0.8256, when the threshold is 0.3 and the depth of the model is 7 layers. However, when k=10, the KMNC of the LeNet model is significantly higher than 0.8256. This indicates that in some cases, the main functional region of the model already contains most neurons. In addition, as shown in Figure \ref{fig:Neuron_Coverages}, the mean score k of the main functional region has an inverse relationship with the coverage of k-multisection neurons. When the model depth is 5 layers, the k-multisection neuron coverage is higher than that of LeNet4, and the coverage of the LeNet5 model lies in between.
As for the answer to question 1, when the model is LeNet, the coverage of k-multisection neurons also decreases with the increase of the model depth, and the rate of decline also increases. When the model is VGG, Figure 6 shows that the k-multisection neuron coverage decreases as the mean score k increases. In addition, the coverage varies less when increasing the depth of the VGG model from 16 to 19 layers. In addition, further observation reveals that when we do not consider the VGG19 model and the value of k ranges from 100 to 5000, the rate of decline of k-multisection neuron coverage increases with the depth of the model. In addition, there is little difference between the k-multisection neuron coverage of all ResNet neural networks. This indicates that when the deep neural network model is deeper than 19 layers and applied to the CIFAR10 dataset, increasing the model depth does not improve the coverage of neurons in the main functional regions.

\noindent\textbf{Answer to RQ2.}
The coverage of KMNC for all deep models decreases as the configuration parameter, the segmentation value k, increases. In addition, when the value of k $\in$ (10,5000), this coverage drop rate increases, and the subsequent drop rate decreases. 

\subsection{Neuron Boundary Coverage}
Boundary region coverage metrics include neuron boundary coverage and strong neuron activation coverage. The neuron boundary coverage metric focuses on how active a neuron is in extreme states, namely edge activity. We use the activity difference to divide the main functional area and the boundary area. Therefore, in order to investigate the relationship between neuronal activity outside the main functional area and different configuration parameters, we mainly analyzed neuronal boundary coverage and strong neuronal activation coverage.

According to the results in Table \ref{tab:NBC_SNAC}, we conducted experiments with different depths of LeNet models under different configuration parameters. Here, $\varepsilon$ represents the distance from the boundary of the main functional area, and $\tau$ represents the size of a single partitioned region in the coverage of k-multiregion neurons. In other words, in the experiment, the neurons boundary scope of coverage concern for ($-\infty$, $L - \varepsilon) \cup (H + \varepsilon, +\infty)$, and strong neuron activation coverage scope of attention $(H + \varepsilon, +\infty)$. In addition, in order to quantitatively analyze the impact of model depth as well as boundary distance on coverage, the experiment in Table \ref{tab:NBC_SNAC} fixed the number of divided regions to 1000. Figure 8 shows that this coverage has different laws for boundary distances $\varepsilon$<0 and $\varepsilon$>0.

\begin{table}[ht]
    \caption{Neuron Boundary Coverage and Strong Neuron Activation Coverage for Deep Models}
    \centering
    \label{tab:NBC_SNAC}
    \begin{tabular}{ll|ccc}
    \toprule
    \multirow{2}{*}{Indicators} & \multirow{2}{*}{Parameters} & \multicolumn{3}{c}{LeNet}  \\*
    \cline{3-5}
        &     & 5 layers     & 6 layers     & 7 layers         \\*
\midrule
    \multirow{8}{*}{NBC}        &    $\varepsilon=-1.5*\tau$     & 0.0714 & 0.082  & 0.0886     \\*
    &    $\varepsilon=-1*\tau$    & 0.066  & 0.0794 & 0.0886   \\*
    &    $\varepsilon=-0.5*\tau$      & 0.0411 & 0.0471 & 0.0486     \\*
    &    $\varepsilon=0*\tau$     & 0.0087 & 0.0066 & 0.006      \\*
    &    $\varepsilon=0.5*\tau$        & 0.0043 & 0.0059 & 0.0056     \\*
    &    $\varepsilon=1*\tau$       & 0.0043 & 0.0059 & 0.0056     \\*
    &    $\varepsilon= 1.5*\tau$        & 0.0043 & 0.0053 & 0.0056     \\*
    &    $\varepsilon=2*\tau$      & 0.0043 & 0.0049 & 0.0056     \\*
\midrule
\multirow{8}{*}{SNAC}       
    &     $\varepsilon=-1.5*\tau$   & 0.2738 & 0.1812 & 0.1182     \\*
    &     $\varepsilon=-1*\tau$    & 0.2024 & 0.1667 & 0.1182     \\*
    &     $\varepsilon=-0.5*\tau$   & 0.1548 & 0.1449 & 0.1008     \\*
    &     $\varepsilon=0*\tau$   & 0.0357 & 0.0543 & 0.062      \\*
    &     $\varepsilon=0.5*\tau$     & 0.0238 & 0.0543 & 0.00581    \\*
    &     $\varepsilon=1*\tau$  & 0.0238 & 0.0543 & 0.0581     \\*
    &     $\varepsilon=1.5*\tau$   & 0.0238 & 0.0507 & 0.0581     \\*
    &     $\varepsilon=2*\tau$   & 0.0238 & 0.0507 & 0.0581   \\ 
\bottomrule
\end{tabular}
\end{table}

\noindent\textbf{Answer to RQ1.}
At boundary distance $\varepsilon$<0, the rate of decrease of the neuron boundary coverage is proportional to the depth of the model. However, when $\varepsilon$>0 and increases infinitely, when the range of the boundary region shrinks, this value has a weak relationship with the model depth and is stable around a certain value.

\noindent\textbf{Answer to RQ2.}
The neuron boundary coverage is high at the boundary distance $\varepsilon$<0 and decreases as the absolute value of the boundary distance shrinks. This also indirectly reflects the effectiveness of coverage detection of neurons in the main functional area. However, when $\varepsilon$>0, the coverage of the neurons at the edge of the model gradually decreases and tends to 0. It is worth noting that the neuron boundary coverage in this range has a similar rate of decline.

\subsection{Strong Neuron Activation Coverage}

\begin{figure}[ht]
\vspace{-4ex}
    \centering
    \subfloat[NBC]{\includegraphics[width=.45\columnwidth]{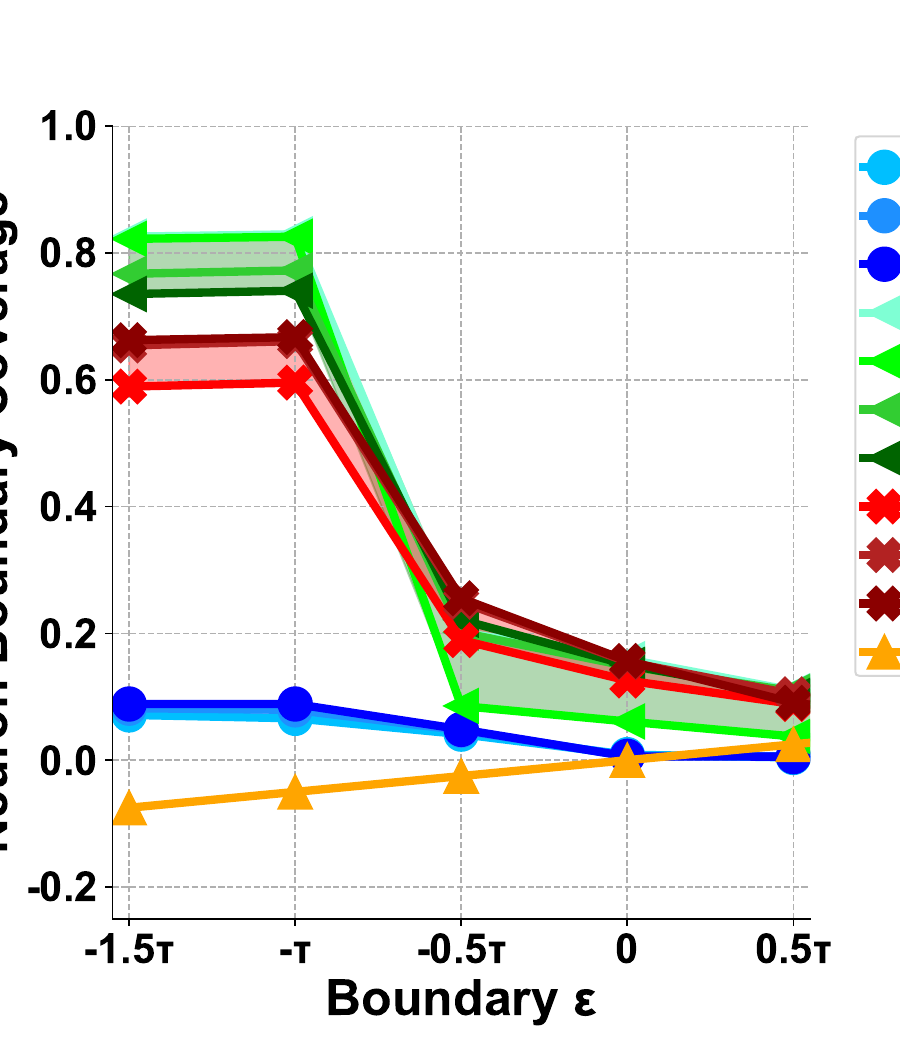}}
    \subfloat[SNAC]{\includegraphics[width=.45\columnwidth]{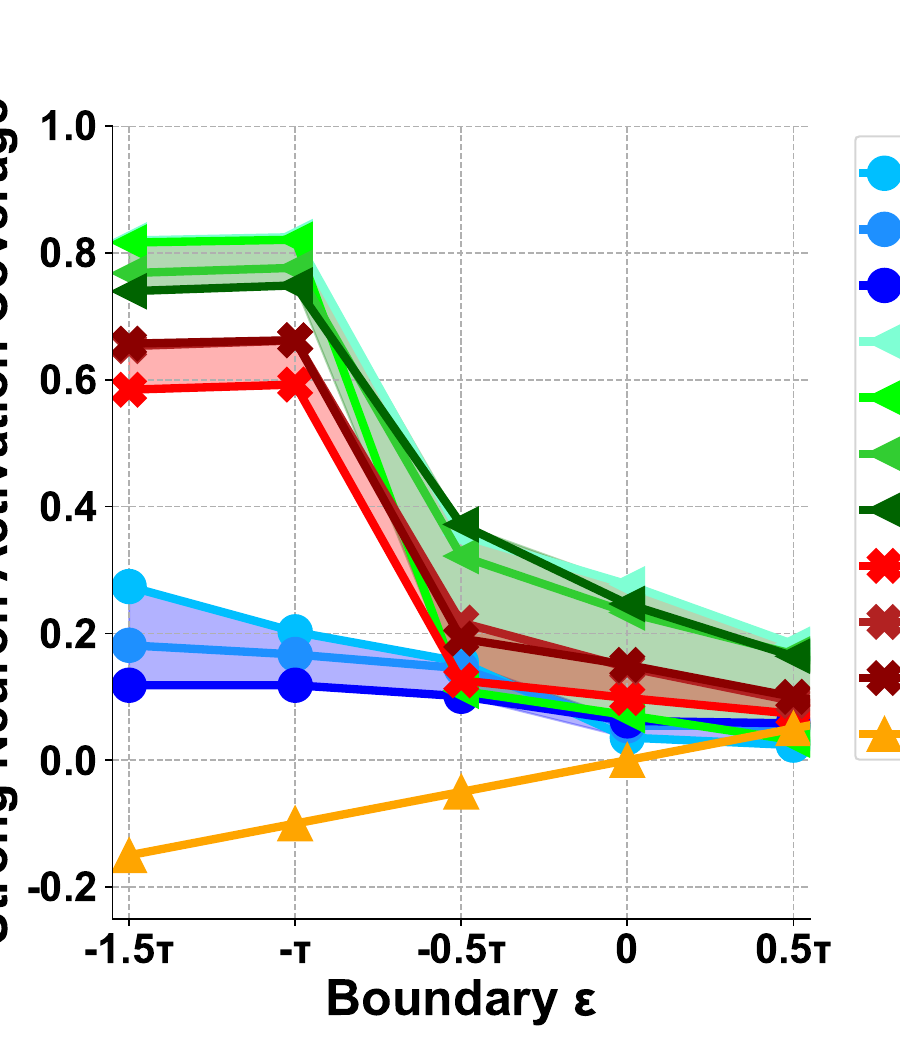}}
    \vspace{-2ex}
    \caption{The NBC and SNAC in Different Models. Blue groups represent the coverages of LeNets, green groups for VGGs, and red groups for ResNets.}
    \label{fig:NBC_SNAC}
\end{figure}

As shown in Figure \ref{fig:NBC_SNAC}, the model depth has an inverse relationship with the initial value of strong neuron activation coverage, and this coverage decreases as the boundary distance increases. When the boundary distance $\varepsilon$<0, that is, for the neurons whose activation range is near the boundary H of the upper domain, the decline rate of the activation coverage of strong neurons slows down with the increase of the depth of the model. Therefore, the neurons with higher activity in the main functional area are more widely distributed in the shallow model. However, when the boundary distance $\varepsilon$>0, the rate of decrease of strong neuron activation coverage approaches 0.

In particular, to quantitatively analyze the neuron boundary coverage of the VGG and ResNet models trained on the CIFAR10 dataset, we fixed the partition region $\tau$ to 10000.

Since the neuron boundary coverage index focuses on neurons in two regions of the upper and lower boundary, that is, hyperactive and inactive neurons, we jointly analyze the two coverage indices in Table \ref{tab:NBC_SNAC} in order to accurately understand the activity rules of strong and weak activity neurons.

\begin{table*}[ht]
    \caption{Top-K Neuron Coverage for Deep Models}
    \vspace{-2ex}
    \centering
    \begin{tabular}{ll|lll|llll|lll}
    \toprule
    \multirow{3}{*}{Indicator}     & \multirow{3}{*}{Parameters} & \multicolumn{10}{c}{Model}\\*
    \cline{3-12}
    &  & \multicolumn{3}{c}{LeNet} & \multicolumn{4}{c}{VGG}    & \multicolumn{3}{c}{ResNet}  \\*
    \cline{3-12}
    &  & 5 layers     & 6 layers     & 7 layers      & 11 layers    & 13 layers   & 16 layers    & 19 layers    & 21 layers    & 37 layers    & 54 layers       \\*
    \midrule
\multirow{7}{*}{Top-K NC} & k=5    & 0.9619 & 0.8304 & 0.8062  & 0.1933 & 0.1923 & 0.1717 & 0.1612 & 0.4004 & 0.3387 & 0.1619    \\*
       & k=10    & 0.8857 & 0.8159 & 0.824   & 0.2657 & 0.2701 & 0.2554 & 0.2384 & 0.5499 & 0.4687 & 0.2359    \\*
          & k=15    & 0.6667 & 0.7667 & 0.823   & 0.3158 & 0.3169 & 0.3155 & 0.2938 & 0.6509 & 0.5698 & 0.2924    \\*
         & k=20     & 0.5    & 0.7163 & 0.8047  & 0.352  & 0.352  & 0.3615 & 0.3404 & 0.7214 & 0.653  & 0.3383    \\*
          & k=25   & 0.4    & 0.6858 & 0.7957  & 0.3834 & 0.3814 & 0.393  & 0.3792 & 0.7865 & 0.724  & 0.3802    \\*
            & k=30    & 0.3333 & 0.6659 & 0.7903  & 0.4084 & 0.4059 & 0.4223 & 0.4108 & 0.8339 & 0.7878 & 0.4181    \\*
           & k=35    & 0.2857 & 0.6482 & 0.787   & 0.4308 & 0.4297 & 0.4483 & 0.4356 & 0.8748 & 0.8338 & 0.4501    \\
    \bottomrule
    \end{tabular}
\end{table*}

When the boundary distance $\varepsilon \in (0, 0.2\tau)$, the values of neuron boundary coverage and strong neuron activation coverage, and their change rates are different in different models. For the VGG model, the value of strong neuron activation coverage is higher than that of neuron boundary coverage, and its decline rate is also positively correlated with the coverage value. This indicates that as the boundary region expands, the number of neurons in the upper boundary decreases. The number of strongly active neurons gradually decreases. Moreover, the number of neurons in the upper boundary is higher than the sum of the number of neurons in the lower and upper boundaries. Thus, within the boundary distance $\varepsilon \in (0, 0.2\tau)$, weakly active neurons are negative in the model.

However, when the boundary distance $\varepsilon$<0, the distribution rules of the neuron boundary coverage and strong neuron activation coverage of the VGG and ResNet models are similar, while there are some slight differences when the boundary distance $\varepsilon$>0. When the boundary distance $\varepsilon$>$0.2\tau$, the VGG model coverage and its change rate tend to be stable from 13 to 19 layers. It is worth noting that when the VGG11 model is within the boundary distance $\varepsilon \in (0.4\tau, 0.5\tau)$, its neuron boundary coverage increases, while the strong neuron activation rate decreases. This means that the number of weakly active neurons in this range increases without affecting the performance of the model.

In the range of boundary distance $\varepsilon \in (-0.1\tau, 0)$ for VGG and ResNet model structures, the decline rate of neuron boundary coverage is high, and it shows a sharp decline trend near $\varepsilon$=0. This phenomenon indicates that the model is close to fit at $\varepsilon$=0, while the neuron activation values vary greatly from $-0.1\tau$ to 0 region, which indicates that this coverage index can be effectively localized to the boundary.

\noindent\textbf{Answer to RQ1.}
In the LeNet model, the reaching value of the strong neuron activation coverage is proportional to the model depth. When the number of model layers is similar, the difference in the reaching value of strong neuron activation coverage decreases with the increase of depth. Assuming $\phi (\alpha, \beta)$ model the coverage of strong neuron activation of $\alpha$ and $\beta$ are absolute value, the $\phi (L (5), L (6)) - \phi (L (6), L (7)) > 0$. Here, $L(\gamma)$ denotes the LeNet model with $\gamma$ layers. Moreover, in the range of boundary distance $\varepsilon \in(-0.1\tau,0)$, the VGG model has the highest rate of decline at depth 13, and the rate of decline is negatively correlated with depth at other depths. However, the neuron boundary coverage of the ResNet model is positively correlated with depth, and the difference in neuron boundary coverage between models with a similar number of layers decreases with the increase of depth. Assume, in other words, $\omega (\alpha, \beta)$ means the model of alpha and beta neurons boundary coverage or absolute difference, the $\omega (R (21), R (37)) - \omega (R (37), R (54)) > 0$. The $R(\gamma)$ represents the ResNet model with $\gamma$ layers.

\noindent\textbf{Answer to RQ2.}
In the LeNet model, the activation coverage of strong neurons decreases with the increase of the threshold boundary distance $\varepsilon$. In addition, its decline rate also increases with the increase of $\varepsilon$, and this coverage and its decline rate stabilize around a fixed value at $\varepsilon$=0.

\subsection{Top-k Neuron Coverage}
Since the three models with different depths of the LeNet structure are trained based on the MNIST dataset, the training and testing process of models such as VGG and ResNet are performed on the CIFAR10 dataset. Therefore, we analyze the coverage of Top-k neurons for LeNet as well as for VGG and ResNet neural network architectures, respectively. Based on the data in Table 5, we summarize multiple Top-k neuron coverage test results in the range of threshold k from 5 to 35.

\begin{figure}[ht]
    \centering
    \includegraphics[width=.45\columnwidth]{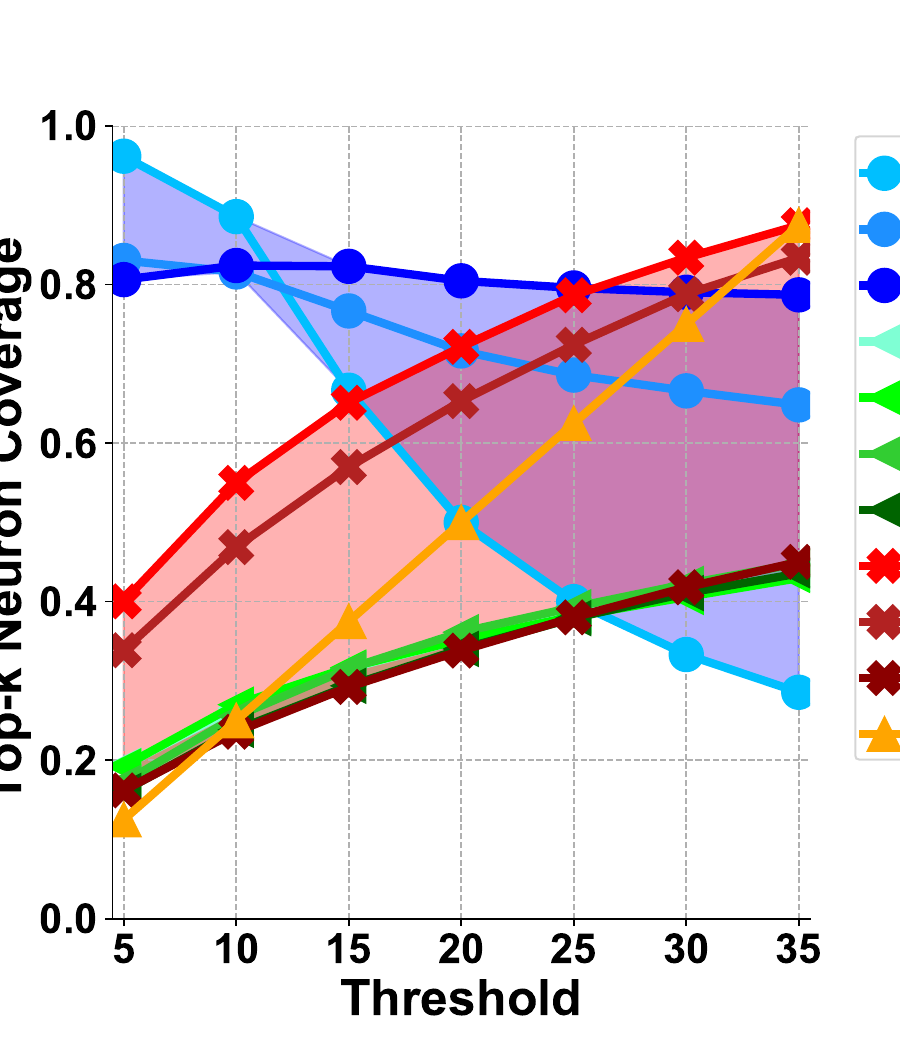}
    \vspace{-2ex}
    \caption{The TopKNC in Different Models. Blue groups represent the coverages of LeNets, green groups for VGGs, and red groups for ResNets.}
    \label{fig:TopkNC}
\end{figure}

\begin{table*}[ht]
    \centering
    \caption{Modified Condition/Decision Coverage for Deep Models}
    \vspace{-2ex}
    \label{tab:mcdc}
    \begin{tabular}{ll|lllllll}
    \toprule
\multirow{2}{*}{Model}  & \multirow{2}{*}{Indicator} & \multicolumn{7}{c}{Dataset Size}              \\*
\cline{3-9}
 &     & 100    & 200    & 400    & 800    & 1600   & 3200   & 6400    \\*
\midrule
\multirow{4}{*}{LeNet (5 layers)} & MC/DC(SS)  & 0.0192 & 0.0337 & 0.0385 & 0.0481 & 0.0481 & 0.0529 & 0.0577  \\*
  & MC/DC(SV)  & 0.2115 & 0.3702 & 0.3846 & 0.4904 & 0.5    & 0.5769 & 0.6346  \\*
   & MC/DC(VS)   & 0.0769 & 0.1346 & 0.1538 & 0.1923 & 0.1923 & 0.2115 & 0.2308  \\*
     & MC/DC(VV)    & 1      & 1      & 1      & 1      & 1      & 1      & 1       \\*
     \midrule
\multirow{4}{*}{LeNet (6 layers)} & MC/DC(SS)   & 0.1443 & 0.1957 & 0.2107 & 0.2823 & 0.2933 & 0.306  & 0.3568  \\*
     & MC/DC(SV)   & 0.082  & 0.1143 & 0.1467 & 0.2003 & 0.2431 & 0.2621 & 0.3031  \\*
     & MC/DC(VS) & 0.5404 & 0.5439 & 0.5439 & 0.5577 & 0.5577 & 0.5577 & 0.5681  \\*
      & MC/DC(VV)  & 0.6686 & 0.6715 & 0.6721 & 0.6721 & 0.6721 & 0.6744 & 0.6744  \\*
      \midrule
\multirow{4}{*}{LeNet (7 layers)} & MC/DC(SS)           & 0.0186 & 0.0409 & 0.051  & 0.0661 & 0.0945 & 0.0969 & 0.1194  \\*
       & MC/DC(SV)   & 0.0125 & 0.0269 & 0.0353 & 0.0454 & 0.06   & 0.0692 & 0.0855  \\*
        & MC/DC(VS)   & 0.1304 & 0.132  & 0.1335 & 0.1339 & 0.1355 & 0.1572 & 0.1531  \\*
        & MC/DC(VV)   & 0.1489 & 0.1503 & 0.1519 & 0.1528 & 0.1532 & 0.1567 & 0.1568 \\
    \bottomrule
    \end{tabular}
    
\end{table*}

\begin{figure*}[ht]
\centering
\subfloat[NC]{\includegraphics[width=.50\columnwidth]{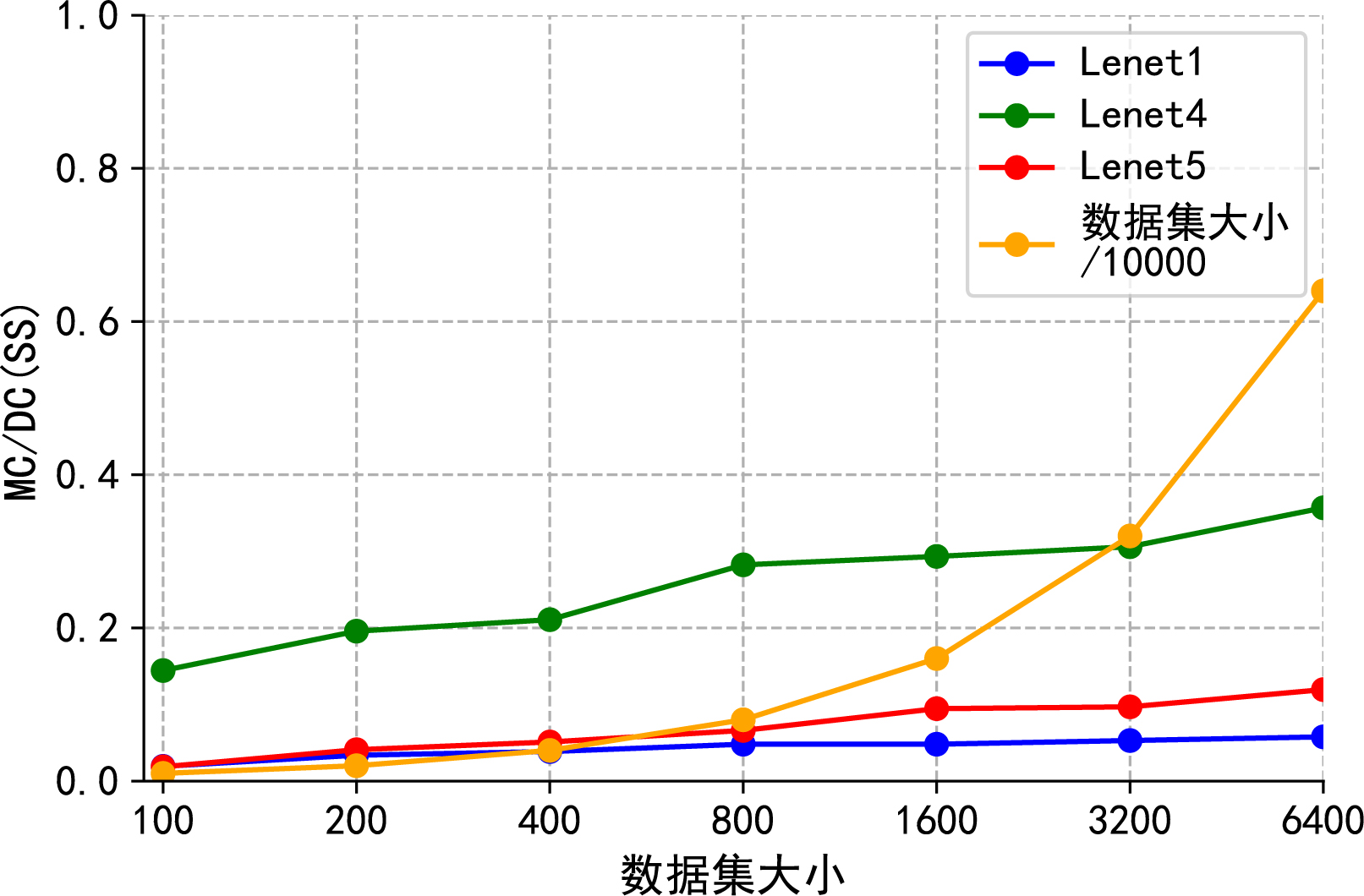}}
\subfloat[KMNC]{\includegraphics[width=.50\columnwidth]{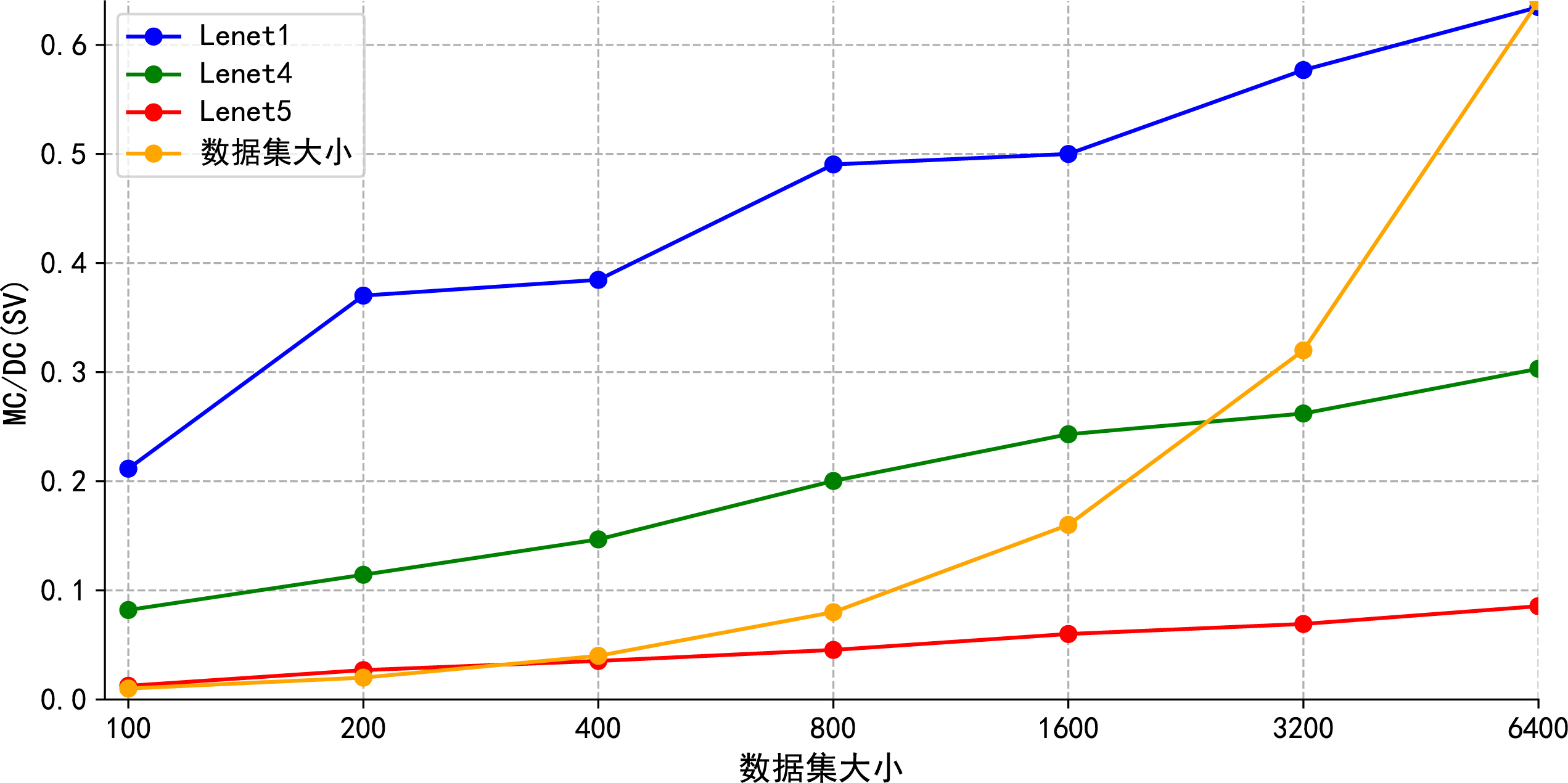}}
\subfloat[NBC]{\includegraphics[width=.50\columnwidth]{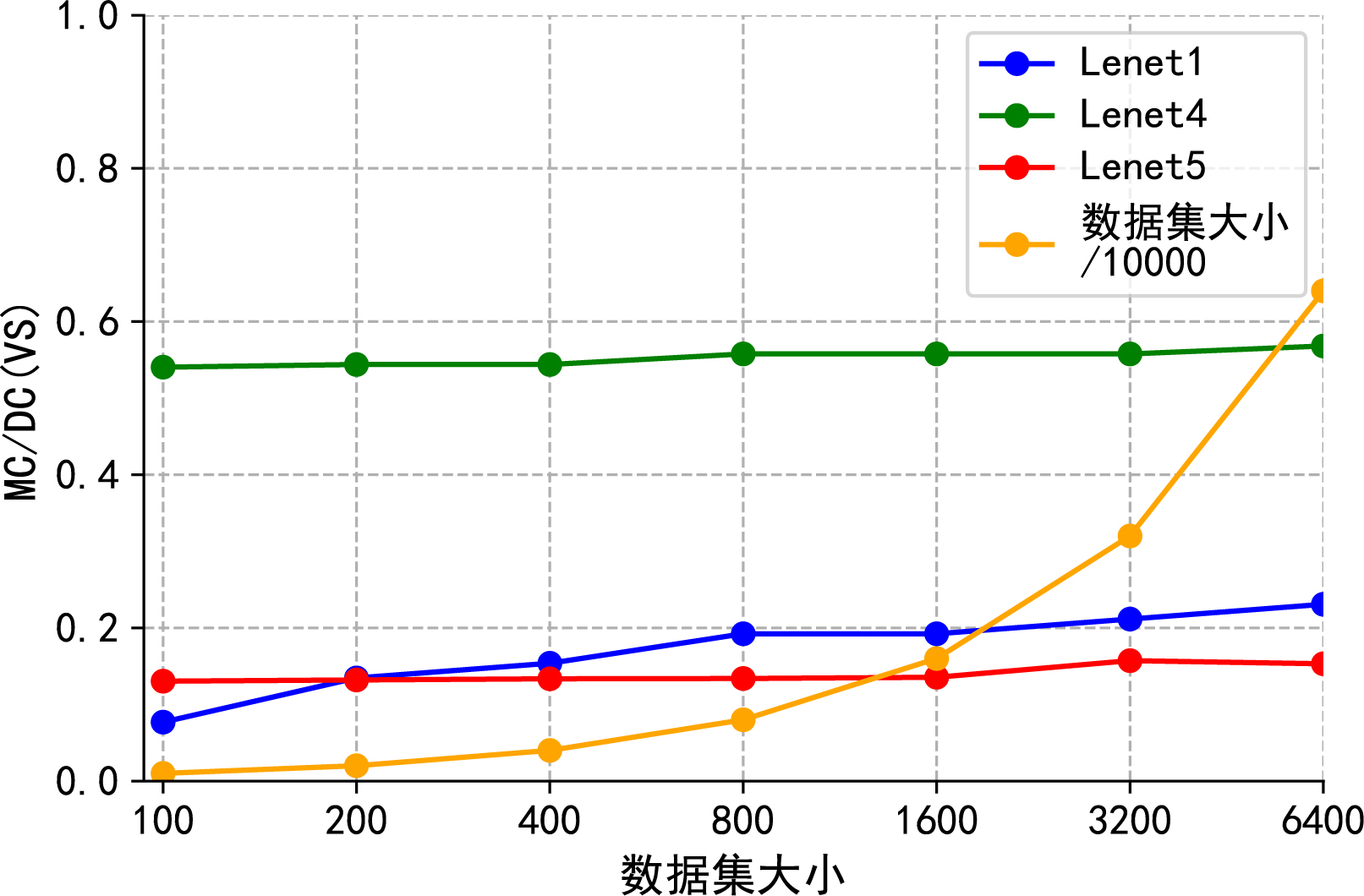}}
\subfloat[SNAC]{\includegraphics[width=.50\columnwidth]{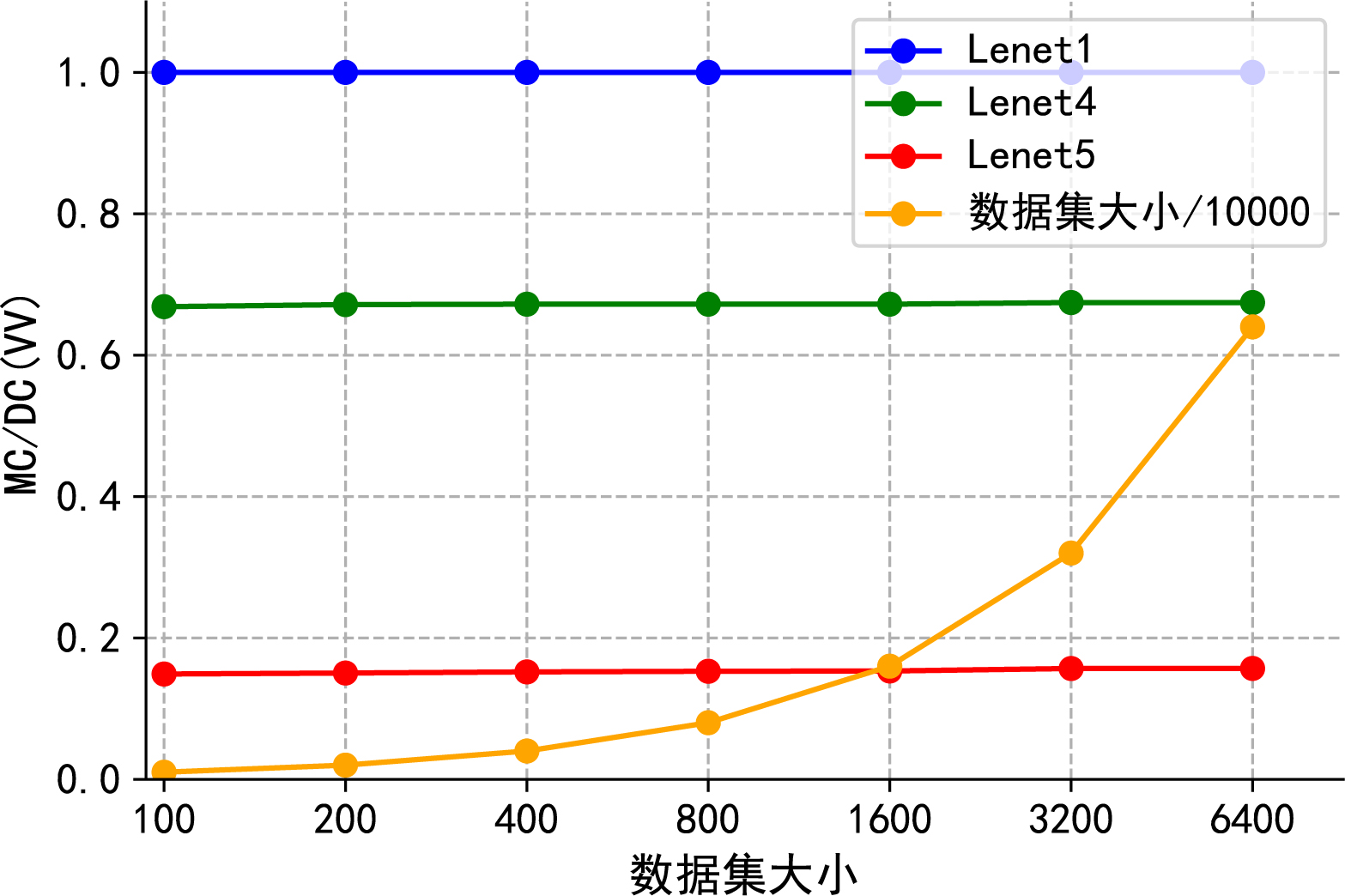}}
\vspace{-2ex}
\caption{The MC/DC Coverage.}
\label{fig:mcdc_Neuron_Coverage}
% \vspace{-4ex}
\end{figure*}

\noindent\textbf{Answer to RQ1.}
For the LeNet model, the rate of Top-k neuron coverage change is negatively correlated with the model depth. Moreover, for the LeNet model with a similar number of layers, the difference in Top-k neuron coverage between them decreases as the depth increases. Assuming $\mu (\alpha, \beta)$ model of alpha and beta Top - k coverage in neurons are absolute value, then we get $\mu (L (5), L (6)) - \mu (L (6), L (7)) > 0$. $L(\gamma)$ represents the LeNet network structure with $\gamma$ layers. Therefore, there is an important relationship between Top-k neuron coverage and model depth in LeNet, a small-scale sequential deep neural network. Top-k neuron coverage counts the total number of strongly active neurons in each layer. However, the number of layers of the model is positively related to the stability of this coverage. Therefore, compared with LeNet1 and LeNet4, the number of strongly active neurons in each layer in LeNet5 is more evenly distributed. In the VGG model, this coverage is weakly correlated with model depth. For the experimental results of the non-sequential neural network structure ResNet in Figure 16, its Top-k neuron coverage is negatively correlated with the model depth, and the Top-k neuron coverage value is higher than the case of the VGG model as a whole. The absolute value of the difference in Top-k neuron coverage of ResNet models with similar layers decreases with decreasing depth, such as $\mu(L(37),L(54))>\mu(L(21),L(37))$.

\noindent\textbf{Answer to RQ2.}
Different deep models of the sequential neural network architecture VGG have similar patterns in the Top-k neuron coverage index. As the threshold k increases, the coverage of Top-k neurons in the VGG model increases, but its change rate gradually decreases. More, as the threshold value increases, the change rate of Top-k neuron coverage gradually slows down and approaches zero.

\subsection{Modified Condition/Decision Coverage}

The MC/DC test index mainly focuses on all the conditions related to the decision; in other words, for each decision, the conditions in its previous layer that can affect the decision. However, to fully analyze these determinations and conditions, we focus on four different coverage methods: Sign-Sign (SS), Sign-Value (SV), Value-Sign (VS), and Value-Value (VV). Here, the sign and the decision represent the function used to detect the change of the activation value of the neuron when the input value changes. We tested this data on different dataset sizes. Also, all other relevant configuration parameters are the same as those in DeepHunter\cite{20}.

\noindent\textbf{Answer to RQ3.} We summarize the four covering method regularities shown in Figure \ref{fig:mcdc_Neuron_Coverage}. In the four coverage methods, even though there are curves with a low change rate, all the coverage rates increase with the increase of the dataset size. Among them, MC/DC(VV) has the highest coverage value and has little relationship with the dataset. In addition, the coverage value is inversely proportional with the model depth, and $\xi (L (5), L (6)) - \xi (L (6), L (7)) > 0$, the $\xi (\alpha, \beta)$ shows two models of $\alpha$ and $\beta$ MC/DC (VV) the difference between the absolute value. In addition, the MC/DC 
(SV) is also inversely proportional to the model depth.

In order to clearly answer the difference between MC/DC and other coverage metrics, we make a comparative analysis from the perspective of model depth, configuration parameters, and change rate. For the model depth, the activation coverage of neurons, strong neurons with boundary distance greater than 0 has a positive correlation law, and the coverage of Top-k neurons, MC/DC (SV), and MC/DC (VV) has a negative correlation law. For the configuration parameters of various indicators, the activation coverage of neurons, Top-k neurons, and strong neurons with boundary distance less than a certain value has a negative correlation law. Strong neuronal activation, MC/DC (SS), MC/DC (SV), MC/DC (VS), and MC/DC (VV) coverage have positive correlation laws for boundary distance greater than 0. In addition, for the rate of change, neurons, strong neuron activation, Top-k neurons, MC/DC (SV), MC/DC (VV) coverage have a negative correlation law with model depth, and neuron boundary coverage has a positive correlation law with model depth.

\section{Future Work}
\label{sec:future_work}

Deep neural network models are widely used in feature extraction for many downstream tasks such as image segmentation, object detection, and pose estimation. With the wide application of deep neural networks, model security testing has also attracted a lot of attention. Therefore, in order to test the security of these models, various coverage metrics have been proposed to detect model units covered by input test cases \cite{5,7,14}. The influencing factors of the coverage detection effect were explored to improve the evaluation accuracy of the test cases on the model. The main process of the coverage index in the model is improved to enhance the interpretability of the deep neural network. In summary, future research on DNN coverage may have the following directions.

\begin{enumerate}
    \item \textbf{Improving the theoretical research system of deep neural network coverage.} At present, the research on neural network coverage index mostly focuses on the design of new neuron coverage index and coverage test priority technology \cite{7,10}. However, for deep neural networks, there is no unified evaluation index for the neuron behavior involved in the coverage index. Although some existing studies analyze the data flow of test input by studying the topology structure in the network. However, there is a lack of relevant neuron behavior analysis to improve the interpretability of the test model. Therefore, it is of great significance to establish an effective analysis framework for neuron coverage and improve the theoretical research system of deep neural network coverage.
    \item \textbf{Exploring the effective combination of various coverage index detection models.} Due to the different emphasis of various coverage indicators, they have different time resource consumption under different configuration parameters. Recent studies have designed a variety of test case priorities to guide different test cases \cite{7,14,22}, and improved the testing effectiveness of neural network models by optimizing the input of test cases. However, not only test cases, but also the selection priority of coverage metrics affect the efficiency of testing work. Therefore, the design of the coverage priority method and the empirical analysis of a variety of coverage combination detection models can provide more efficient model testing work, and also help to explore testing strategies from more perspectives.
    \item  \textbf{Designing an effective test system for different structural models.} Although the sequential network structure model and the non-sequential network structure model have the same regular relationship with their configuration parameters and model depth when facing the input samples, there are obvious differences in the layer index Top-k neuron coverage. Therefore, it is of great significance to design a universal structural coverage index for different structural models to break through the structural test of deep neural networks with complex structures.
\end{enumerate}

\section{Conclusion}
\label{sec::conclusion}
This paper presents an empirical study of coverage metrics in deep neural networks. First, we illustrate the coverage metric and the traditional coverage metric MC/DC. Then, we point out the relationship between model depth, configuration parameters with respect to different coverage metrics, and the regular relationship at different configuration cases. To answer these questions, we conduct extensive empirical experiments on 6 different coverage rates in 10 deep neural models with different number of layers to investigate the properties of various coverage rates. Finally, we provide the future research directions of coverage metrics in deep neural networks, hoping that we can provide comprehensive cognition and reference for researchers interested in coverage work, and make further contributions to the security testing of DNNs.

\normalem
\bibliographystyle{ACM-Reference-Format}

\bibliography{ref}

\end{document}